# Antisemitic Messages? A Guide to High-Quality Annotation and a Labeled Dataset of Tweets

Gunther Jikeli,[1] Sameer Karali,[1] Daniel Miehling,[2] and Katharina Soemer[1]

Indiana University Bloomington, USA,[1,] Technical University Berlin, Germany[2]

{gjikeli, skarali, damieh, ksoemer}@iu.edu

**Abstract**

One of the major challenges in automatic hate speech detection is the lack of datasets that cover a wide range of biased and unbiased messages and that are consistently labeled. We propose a labeling procedure that addresses some of the common weaknesses of labeled datasets.

We focus on antisemitic speech on Twitter and create a labeled dataset of 6,941 tweets that cover a wide range of topics common in conversations about Jews, Israel, and antisemitism between January 2019 and December 2021 by drawing from representative samples with relevant keywords.

Our annotation process aims to strictly apply a commonly used definition of antisemitism by forcing annotators to specify which part of the definition applies, and by giving them the option to personally disagree with the definition on a case-by-case basis. Labeling tweets that call out antisemitism, report antisemitism, or are otherwise related to antisemitism (such as the Holocaust) but are not actually antisemitic can help reduce false positives in automated detection.

The dataset includes 1,250 tweets (18%) that are antisemitic according to the International Holocaust Remembrance Alliance (IHRA) definition of antisemitism.

It is important to note, however, that the dataset is not comprehensive. Many topics are still not covered, and it only includes tweets collected from Twitter between January 2019 and December 2021. Additionally, the dataset only includes tweets that were written in English. Despite these limitations, we hope that this is a meaningful contribution to improving the automated detection of antisemitic speech.

## Introduction and Related Work

Automated hate speech detection has advanced significantly in recent years with the development of deep learning techniques and large-scale training data. Many studies have shown that machine learning algorithms, especially deep neural networks, can achieve high accuracy in detecting hate speech within test datasets.

Hate speech detection using automated methods such as BERT, ELECTRA, or Perspective API has been developed to investigate large datasets containing toxic speech patterns and conspiracy-related content (Alkomah and Ma 2022; MacAvaney et al. 2019; Poletto et al. 2021).

However, the task of detecting hate speech remains challenging for several reasons (Yin and Zubiaga 2021). First, the datasets on which the models are trained are relatively small. They do not include all variations of hate speech manifestations in a rapidly changing environment. Second, there is no single definition of hate speech. It can be expressed in a variety of ways, often involves a high degree of subjectivity, and depends on cultural, social, and historical factors, making it difficult to identify and classify consistently. Third, proper classification often requires more context than what is readily available, such as previous discussions in a thread or a history of ironic messages from particular users. Calling out hate speech or reporting on stereotypes often results in false positives. A quick test with ChatGPT can illustrate this. ChatGPT correctly identifies antisemitic stereotypes in the message *"Fox News trashes Georges Soros while praising Joe Rogan using some antisemitic tropes – puppet master using his money to control the world. Then Pete Hegseth goes into a rant about the nonsense conspiracy theory Cultural Marxism. This is from Fox & Friends morning show."* However, ChatGPT classifies it as an antisemitic message itself.

Despite these challenges, the research community has made significant progress in developing models that can detect hate speech. These models often rely on a combination of linguistic features, such as word n-grams, word embeddings, and contextual features, such as the identity of the author or the presence of certain keywords.

It's worth noting that while the quality of these models has improved, they are not perfect and can still produce false positives or false negatives. As a result, it is important to use these models as part of a larger framework that includes human review and oversight. It is also important to ensure that these models are trained on diverse, representative data and that their results are interpretable and transparent.

Narrowing hate speech to hostile attitudes toward specific communities or groups of people can help improve automated detection models, as it may be easier to consistently label datasets with a precise definition.

While there are many labeled datasets on hate speech in general, few datasets focus specifically on antisemitism. (Chandra et al. 2021) built a labeled dataset on antisemitism from 3,102 posts on Twitter and 3,509 posts on Gab, focusing on messages containing both images and text, as well as words related to Jews, such as "Jewish," "Hasidic," "Hebrew," "Semitic," "Judaistic," "israeli," "yahudi," "yehudi," and also slurs. The Twitter dataset, including IDs and annotations, is publicly available.[1] Three annotators labeled posts as antisemitic or not and classified antisemitic posts into one of the four categories: political, economic, religious, or racial antisemitism. (Steffen et al. 2022) labeled a dataset of 3,663 German-language Telegram messages about antisemitism and conspiracy theories. They retrieved the messages from Telegram channels that were used to protest government measures to contain the pandemic. The messages were posted between March 11, 2020, and December 19, 2021. Both projects make their datasets available publicly or, in the latter case, on request. Both projects use the International Holocaust Remembrance Alliance's Working Definition of Antisemitism (IHRA Definition) as a guideline for determining whether a message is antisemitic or not.[2] This is also the case for Schwarz-Friesel's comprehensive study of online messages in German (Schwarz-Friesel 2019). Guhl et al.'s study of the German far-right (Guhl, Ebner, and Rau 2020), and the ongoing Decoding Antisemitism project, which examines online comments on articles in mainstream media outlets in English, German, and French (Ascone et al. 2022; Becker and Allington 2021). We have shown that the definition can be successfully used to classify online messages when inferences from the definition, such as "classical antisemitic stereotypes," are spelled out (Jikeli, Cavar, and Miehling 2019).

Research on automated detection of hate speech and conspiracy theories related to the Covid-19 pandemic or QAnon has shown that Jews figure prominently in conspiracy fantasies (Vergani et al. 2022; Hoseini et al. 2021; La Morgia et al. 2021). Automated detection of antisemitic speech could also help to identify conspiracy theories.

However, time-consuming manual annotation and consistent labeling are the bottleneck for most supervised machine learning projects. Our project on antisemitic tweets is in principle not different from many other projects on hate speech dataset, which include defining a classification scheme, labeling guidelines, collecting adequate data, preprocessing this data according to the task, training experts on labeling, and building a final corpus (Pustejovsky and Stubbs 2013).

However, we propose a number of measures to ensure the production of high quality datasets. We are making the IDs and the text of our dataset available, along with our label of whether or not they are antisemitic. Later this year, we will add the label of calling out antisemitism. The usernames in the tweets are not anonymized, as we believe this information may be useful for further research.

## Generating Our Corpus

Our corpus covers a wide range of antisemitic and non-antisemitic conversations about Jews on Twitter from 2019 to 2021. We used Indiana University's Observatory on Social Media (OSoMe) database to identify relevant messages. The OSoMe database contains 10 percent of all live tweets on a statistically relevant basis. We queried with two keywords that are likely to result in a wide range of conversations about Jews as a religious, ethnic, or political community: "Jews" and "Israel." We then added samples with more targeted keywords that are likely to generate a high percentage of antisemitic tweets, namely the slurs

---

[1] https://github.com/mohit3011/Online-Antisemitism-Detection-Using-MultimodalDeep-Learning

[2] IHRA Working Definition of Antisemitism, see https://www.holocaustremembrance.com/resources/working-definitions-charters/working-definition-antisemitism

"K---s" [3] and "ZioNazi*." We ran 14 queries for different time periods between January 2019 and December 2021. The queries returned tweet ID numbers. We then randomly sampled 2,000 tweets from each query, pulled the text and metadata from the Twitter archive from those that were still live, filtered for English tweets using Google's language detection library, and randomly selected 500 tweets from the remaining tweets.[4] We took screenshots of the sampled tweets for documentation. We repeated this process for four queries, resulting in two samples for those queries and 18 samples in total, see Table 1.

## Annotation

We annotated the tweets, considering the text, images, videos, and links, in their "natural" context, including threads. We used a detailed annotation guideline (Jikeli, Cavar, and Miehling 2019), based on the IHRA Definition, which has been endorsed and recommended by more than 30 governments and international organizations[5] and is frequently used to monitor and record antisemitic incidents. We divided the definition into 12 paragraphs. Each of the paragraphs addresses different forms and tropes of antisemitism. We created an online annotation tool (https://annotationportal.com) to make labeling easier, more consistent, and less prone to errors, including in the process of recording the annotations. The portal displays the tweet and a clickable annotation form, see Figure 1. It automatically saves each annotation, including the time spent labeling each tweet.

The Annotation Portal retrieves live tweets by referencing their ID number. Our annotators first look at the tweet, and if they are unsure of the meaning, they are prompted to look at the entire thread, replies, likes, links, and comments. A click on the visualized tweet opens a new tab in the browser, displaying the message on the Twitter page in its "natural" environment.

The portal is designed to help annotators consistently label messages as antisemitic or not according to the IHRA definition. After verifying that the message is still live and in English, they select from a drop-down menu where they classify the message as "confident antisemitic," "probably antisemitic," "probably not antisemitic," "confident not antisemitic," or "don't know." The annotation guideline, including the definition, is linked in a PDF document.

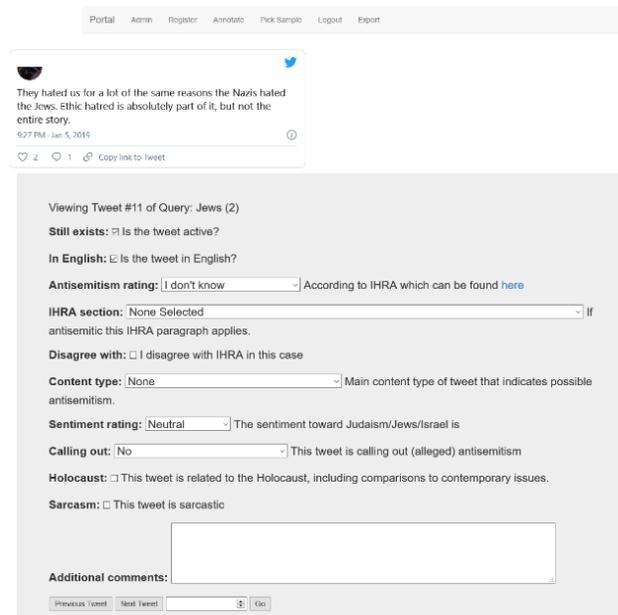

Figure 1: Annotation Portal with tweet example[6]

All annotators are familiar with the definition and have been trained on test samples. They have also taken at least one academic course on antisemitism or have done research on antisemitism. We consider them to be expert annotators. Eight such expert annotators of different religions and genders labeled the 18 samples, two for each sample in alternating configurations.

When annotators label a tweet as "probably" or "confident" antisemitic, they must also select an applicable section of the IHRA definition to move on to the next tweet. If annotators feel the tweet is antisemitic, but no section of the definition applies, they will classify the tweet as not antisemitic according to the IHRA definition, and check a box indicating that they disagree with the IHRA definition for that tweet. Annotators can also use this the other way around, that is, they can label the message as antisemitic according to the IHRA definition, but personally disagree with it in a particular case. The option to personally disagree with the definition on a case-by-case basis is intended to encourage a stricter application of the IHRA definition rather than the individual definitions of the annotators.

---

[3] "K----s" stands for the antisemitic slur "kikes."
[4] For the queries with the two slurs the sample size was smaller because fewer tweets remained after this process.
[5] https://www.holocaustremembrance.com/resources/working-definitions-charters/working-definition-antisemitism/adoption-endorsement
[6] This is a screenshot of our updated form. The question about the content type was not used for the annotation of all samples in this dataset.

Asking annotators to choose between a very negative, negative, neutral, positive, or very positive sentiment for the tweet regarding Jews, Judaism, or Israel further helps annotators apply the definition by allowing them to express that the tweet has a negative sentiment even if they are unable to find a part of the definition that applies. Messages that call out or report antisemitism are also flagged, as are tweets that are sarcastic and tweets that relate to the Holocaust, including comparisons to contemporary issues. Since the Covid pandemic, we have seen an increase in Holocaust distortions, most of which do not fall under the IHRA definition. We have therefore added a Holocaust distortion label.

As an element of what we consider annotation reliability, our annotators meet on a weekly basis to discuss potentially difficult tweets. A tweet can be considered difficult if its content or context is not easily understandable, or if it is unclear which paragraph of the IHRA definition applies.

Two annotators labeled each sample. After both annotators completed their annotations, they discussed their disagreements about whether or not the tweets were antisemitic.[7] Reasons for disagreement included mislabeling due to fatigue, lack of understanding of the context, and overlooking some aspects of the messages. In almost all cases, the discussion led to an agreement. The tweets that did not reach agreement, that is, where one annotator labeled the message as antisemitic and the other did not, were not included in our final labeled dataset. Table 1 shows the annotation results after discussion with the remaining number of tweets and the percentage of antisemitic tweets.[8]

Many dataset labeling projects provide kappa coefficients to measure the quality of inter-annotator agreement. This does not make sense in our case because we discuss all disagreements, and few disagreements remain. However, kappa requires independent classification. Therefore, kappa is an artificial value for our dataset. It is very close to 1. Our overall pre-discussion Cohen's kappa is 0.66, varying from sample to sample.[9] Not surprisingly, it is lower when the data has a skewed distribution, that is, when either very few tweets are antisemitic or if very few tweets in a sample are not antisemitic. Kappa does not seem appropriate for measuring annotation reliability for our dataset, and perhaps for social data annotation in general.

The weekly group discussions and the discussions among the annotators helped the annotators to better understand the context of online conversations about events and online celebrities in the US, UK, India, or elsewhere. The annotators became increasingly familiar with the contexts because they often revolved around similar topics.

**Annotation Results**

In our published labeled dataset, we use binary categories, treating ratings of confident/probably not antisemitic, and don't know as not antisemitic and probably/confident antisemitic as antisemitic. While annotators discussed disagreements about their antisemitism ratings, they did not discuss disagreements about their "calling out" ratings. The label "calling out" was more common than the label "antisemitism" in many samples, but it was inconsistent, especially in the annotations at the beginning of the project. We are relabeling the dataset for "calling out" and will publish the results in an update of the dataset. An overview of the annotation results of the label "antisemitism" per sample can be found in Table 1.

| Sample | Keyword | Timespan | Number of tweets in published labeled dataset | Percentage of antisemitic tweets |
|---|---|---|---|---|
| 1 | Jews | Jan.-Dec. 2019 | 432 | 6.3 % |
| 2 | Jews | Jan.-Dec. 2019 | 409 | 7.6 % |
| 3 | Jews | Jan.-Apr. 2020 | 460 | 12.2 % |
| 4 | Jews | Jan.-Apr. 2020 | 423 | 11.6 % |
| 5 | Jews | May-Aug. 2020 | 390 | 14.1 % |
| 6 | Jews | May-Aug. 2020 | 387 | 16.3 % |
| 7 | Jews | Sep.-Dec. 2020 | 402 | 10 % |

---

[7] Annotators discussed messages that one of them labeled as confident or probably antisemitic and the other one labeled as confident or probably not antisemitic or "I don't know."
[8] Before publishing the dataset, we checked for errors, including a possible overlap between tweets labeled as antisemitic and tweets calling out antisemitism. They were relabeled, resulting in the correction of 9 tweets.
[9] Cohen's kappa was calculated for dichotomized values, "I don't know" falls under "not antisemitic."

| | | | | |
|---|---|---|---|---|
| 8 | Jews | Sep.-Dec. 2020 | 390 | 7.4 % |
| 9 | Jews | Jan.-Apr. 2021 | 453 | 6 % |
| 10 | Jews | May-Aug. 2021 | 414 | 19.6 % |
| 11 | Jews | Sep.-Dec. 2021 | 445 | 5.6 % |
| 12 | Israel | Jan.-Apr. 2020 | 294 | 11.6 % |
| 13 | Israel | May-Aug. 2020 | 408 | 13.7 % |
| 14 | Israel | Sep.-Dec. 2020 | 408 | 15.2 % |
| 15 | Israel | Jan.-Apr. 2021 | 414 | 12.3 % |
| 16 | ZioNazi* | Jan.-Dec. 2019 | 375 | 89.3 % |
| 17 | ZioNazi* | Jan.-Apr. 2020 | 154 | 85.7 % |
| 18 | kikes | Jan.-Dec. 2019 | 283 | 34.3 % |
| | | Jan. 2019 - Dec. 2021 | **6941** | **18** % |

Table 1: Overview of samples in dataset

Our Gold Standard includes 6,941 tweets with keywords related to antisemitism and Jewish life, of which 1,250 tweets (18%) are antisemitic according to the IHRA definition. 1,499 tweets (22%) were sent in 2019, 3,716 tweets (54%) in 2020 and 1726 tweets (25%) are from 2021. Out of the 6,941 tweets, 4,605 tweets (66%) are from queries with the keyword "Jews," 1,524 tweets (22%) with the keyword "Israel," 529 tweets (8%) with the slur "ZioNazi*" and 283 tweets (4%) with the slur "K---s." Some of the keywords may also appear in other samples, e.g., a tweet may contain both the word "Jews" and "Israel."

Out of 4,605 tweets containing the keyword "Jews," 483 tweets (11%) are considered antisemitic. Out of 1,524 tweets containing the keyword "Israel," 203 tweets (13%) are antisemitic. Our dataset contains 283 tweets with the antisemitic slur "k---s." It is not surprising that many messages, 34% in our samples (97 messages), with this keyword are antisemitic, but we also noticed that many tweets containing the slur "k---s" are calling out the use of this term. In contrast, the vast majority of tweets with the slur "ZioNazi*" are antisemitic: 467 out of 529, or 88%.

The labeled dataset on antisemitism is now awaiting testing, and to facilitate this we have made it available on Zenodo (Jikeli et al. 2023).

## Discussion

Our labeled dataset of 6,941 tweets is based on representative samples of tweets containing the common keywords "Jews" and "Israel" and keywords more likely to be used in antisemitic contexts, the slurs "ZioNazi*" and "k---s." It includes 1,250 tweets (18%) that are antisemitic according to IHRA's Working Definition of Antisemitism.

The majority of the tweets (66%) come from queries with the keyword "Jews," which is representative of a continuous time period from January 2019 to December 2021. It is reasonable to assume that our dataset is a good reflection of discussions on Twitter about Jews and covers the most prevalent topics, at least when the word "Jews" is mentioned and for the three-year period covered by the dataset. 483 tweets (11%) with the keyword "Jews" were labeled as antisemitic. It is also reasonable to assume that they cover most of the relevant topics of antisemitic discussions about Jews on Twitter during this period.

203 out of 1,524 (13%) tweets about Israel are antisemitic. They are likely to cover the main tropes and discussions during this period. However, this period does not include heightened tensions in the Israeli-Palestinian conflict, such as the flare-up of the conflict in May 2021. We will be updating our dataset with data from all of 2021 and 2022 in the near future.

The slurs "ZioNazi*" and "K---s" are not used very often by Twitter users compared to the words "Jews" and "Israel." While there are millions of tweets containing the latter words each year, there are "only" a few tens of thousands of tweets containing these slurs. The majority of tweets containing the slur "Zionazi*," are used in an approving way and have an antisemitic message (88%). This is not the case when the word "k---s" is used. Only one-third are antisemitic.

The tweets in our dataset cover a wide range of antisemitic and non-antisemitic conversations about Jews. However, the dataset needs to be enlarged and constantly updated to cover all topics comprehensively. Labeling tweets that call out antisemitism, report antisemitism, or are otherwise related to antisemitism (such as the Holocaust) but are not actually antisemitic can help reduce false positives in automated detection.

It is important to annotate online messages in their "natural" context. Context, including pictures, memes, or previous comments within a thread, can completely change the meaning of a message.

The annotation process encourages annotators to apply a widely used definition of antisemitism consistently, even if they disagree on certain aspects or on certain cases. Our annotation process appears to be robust although it is difficult to measure because a key element of our annotation process is the discussion among annotators of their disagreements and weekly discussions of tweets that are difficult to classify. This violates the assumption of independent classification in kappa calculations. Percentage agreement and Cohen's kappa almost reach their maximum after annotators discuss and revisit tweets on which they previously disagreed. In the pre-discussion annotation, we do not aim for 100% agreement. Rather, we want annotators from different perspectives to fully understand the messages of each tweet, which they can then explain in discussions focused on deciding whether or not it is antisemitic according to the IHRA definition. The dataset only includes tweets with 100% agreement between annotators. The pre-discussion inter-rater reliability has a kappa value of 0.66. We consider the training of qualified annotators and the discussion process to be essential for building an accurately labeled dataset.

## Acknowledgements

This work used Jetstream2 at Indiana University through allocation HUM200003 from the Advanced Cyberinfrastructure Coordination Ecosystem: Services & Support (ACCESS) program, which is supported by National Science Foundation grants #2138259, #2138286, #2138307, #2137603, and #2138296.
We are grateful for the support of Indiana University's Observatory on Social Media (OSoMe) (Davis et al. 2016) and the contributions and annotations of all team members in our Social Media & Hate Research Lab at Indiana University's Institute for the Study of Contemporary Antisemitism, especially Grace Bland, Elisha S. Breton, Kathryn Cooper, Robin Forstenhäusler, Sophie von Máriássy, Mabel Poindexter, Jenna Solomon, Clara Schilling, and Victor Tschiskale.

## References

Alkomah, Fatimah, and Xiaogang Ma. 2022. "A Literature Review of Textual Hate Speech Detection Methods and Datasets." *Information* 13 (6): 273. https://doi.org/10.3390/info13060273.

Ascone, Laura, Matthias J. Becker, Matthew Bolton, Alexis Chapelan, Jan Krasni, Karolina Placzynta, Marcus Scheiber, Hagen Troschke, and Chloé Vincent. 2022. "Decoding Antisemitism: An AI-Driven Study on Hate Speech & Imagery Online. Third Discourse Report," April. https://doi.org/10.14279/DEPOSITONCE-15314.

Becker, Matthias J., and Daniel Allington. 2021. "Decoding Antisemitism: An AI-Driven Study on Hate Speech and Imagery Online. Discourse Report No. 2. Technical Report." Berlin: Technical University of Berlin Centre for Research on Antisemitism.

Chandra, Mohit, Dheeraj Pailla, Himanshu Bhatia, Aadilmehdi Sanchawala, Manish Gupta, Manish Shrivastava, and Ponnurangam Kumaraguru. 2021. "'Subverting the Jewtocracy': Online Antisemitism Detection Using Multimodal Deep Learning." *ArXiv:2104.05947 [Cs]*, April. http://arxiv.org/abs/2104.05947.

Davis, Clayton A., Giovanni Luca Ciampaglia, Luca Maria Aiello, Keychul Chung, Michael D. Conover, Emilio Ferrara, Alessandro Flammini, et al. 2016. "OSoMe: The IUNI Observatory on Social Media." *PeerJ Computer Science* 2 (October): e87. https://doi.org/10.7717/peerj-cs.87.

Guhl, Jakob, Julia Ebner, and Jan Rau. 2020. "The Online Ecosystem of the German Far-Right." London, Washington DC, Beirut, Toronto: Institute for Strategic Dialogue. https://www.bosch-stiftung.de/en/publication/online-ecosystem-german-far-right.

Hoseini, Mohamad, Philipe Melo, Fabricio Benevenuto, Anja Feldmann, and Savvas Zannettou. 2021. "On the Globalization of the QAnon Conspiracy Theory Through Telegram." https://doi.org/10.48550/ARXIV.2105.13020.

Jikeli, Gunther, Damir Cavar, and Daniel Miehling. 2019. "Annotating Antisemitic Online Content. Towards an Applicable Definition of Antisemitism." *ArXiv:1910.01214 [Cs.CY]*, arXiv preprint, . https://doi.org/10.5967/3r3m-na89.

Jikeli, Gunther, Sameer Karali, Daniel Miehling, and Katharina Soemer. 2023. "Antisemitism on Twitter: A Dataset for Machine Learning and Text Analytics." Zenodo. https://doi.org/10.5281/ZENODO.7872835.

La Morgia, Massimo, Alessandro Mei, Alberto Maria Mongardini, and Jie Wu. 2021. "Uncovering the Dark Side of Telegram: Fakes, Clones, Scams, and Conspiracy Movements." https://doi.org/10.48550/ARXIV.2111.13530.

MacAvaney, Sean, Hao-Ren Yao, Eugene Yang, Katina Russell, Nazli Goharian, and Ophir Frieder. 2019. "Hate Speech Detection: Challenges and Solutions." Edited by Minlie Huang. *PLOS ONE*

14 (8): e0221152. https://doi.org/10.1371/journal.pone.0221152.

Poletto, Fabio, Valerio Basile, Manuela Sanguinetti, Cristina Bosco, and Viviana Patti. 2021. "Resources and Benchmark Corpora for Hate Speech Detection: A Systematic Review." *Language Resources and Evaluation* 55 (2): 477–523. https://doi.org/10.1007/s10579-020-09502-8.

Pustejovsky, J., and Amber Stubbs. 2013. *Natural Language Annotation for Machine Learning*. Sebastopol, CA: O'Reilly Media.

Schwarz-Friesel, Monika. 2019. *Judenhass im Internet Antisemitismus als kulturelle Konstante und kollektives Gefühl*. Berlin; Leipzig: Hentrich & Hentrich.

Steffen, Elisabeth, Helena Mihaljević, Milena Pustet, Nyco Bischoff, María do Mar Castro Varela, Yener Bayramoğlu, and Bahar Oghalai. 2022. "Codes, Patterns and Shapes of Contemporary Online Antisemitism and Conspiracy Narratives -- an Annotation Guide and Labeled German-Language Dataset in the Context of COVID-19." https://doi.org/10.48550/ARXIV.2210.07934.

Vergani, Matteo, Alfonso Martinez Arranz, Ryan Scrivens, and Liliana Orellana. 2022. "Hate Speech in a Telegram Conspiracy Channel During the First Year of the COVID-19 Pandemic." *Social Media + Society* 8 (4): 205630512211387. https://doi.org/10.1177/20563051221138758.

Yin, Wenjie, and Arkaitz Zubiaga. 2021. "Towards Generalisable Hate Speech Detection: A Review on Obstacles and Solutions." *PeerJ Computer Science* 7 (June): e598. https://doi.org/10.7717/peerj-cs.598.